\documentclass[10pt,twocolumn,letterpaper]{article}

\usepackage{iccv}
\usepackage{times}
\usepackage{epsfig}
\usepackage{graphicx}
\usepackage{amsmath}
\usepackage{amssymb}
\usepackage{multicol}
\usepackage{multirow}
\usepackage{subcaption}
\usepackage[normalem]{ulem}
\usepackage[super]{nth}
\usepackage{cite}

\usepackage[pagebackref=true,breaklinks=true,letterpaper=true,colorlinks,bookmarks=false]{hyperref}

\iccvfinalcopy 

\def\iccvPaperID{48} 

\ificcvfinal\fi

\begin{document}

\title{On the Effectiveness of LayerNorm Tuning for \\Continual Learning in Vision Transformers}

\author{\vspace{0.5em}
\setlength\tabcolsep{0.5em}
\begin{tabular}{ccccc} 
Thomas De Min$^{1}$ & Massimiliano Mancini$^{1}$ &  Karteek Alahari$^{2}$ & Xavier Alameda-Pineda$^{2}$  & Elisa Ricci$^{1,3}$ \tabularnewline
\end{tabular}
\\
\renewcommand{\arraystretch}{0.5}
\begin{tabular}{ccc} 
    $^1$University of Trento & $^2$Inria$^\dagger$& $^3$Fondazione Bruno Kessler
\end{tabular}}
\maketitle
\def\thefootnote{$\dagger$}\footnotetext{\scriptsize{Univ.\ Grenoble Alpes, CNRS, Grenoble INP, LJK, 38000 Grenoble, France.}}
\ificcvfinal\thispagestyle{empty}\fi

\renewcommand\thefootnote{\arabic{footnote}}

\begin{abstract}
  State-of-the-art rehearsal-free continual learning methods exploit the peculiarities of Vision Transformers to learn task-specific prompts, drastically reducing catastrophic forgetting. However, there is a tradeoff between the number of learned parameters and the performance, making such models computationally expensive. In this work, we aim to reduce this cost while maintaining competitive performance. We achieve this by revisiting and extending a simple transfer learning idea: learning task-specific normalization layers. Specifically, we tune the scale and bias parameters of LayerNorm for each continual learning task, selecting them at inference time based on the similarity between task-specific keys and the output of the pre-trained model. To make the classifier robust to incorrect selection of parameters during inference, we introduce a two-stage training procedure, where we first optimize the task-specific parameters and then train the classifier with the same selection procedure of the inference time. Experiments on ImageNet-R and CIFAR-100 show that our method achieves results that are either superior or on par with {the state of the art} while being computationally cheaper.\footnote{Code is available at
  {\url{https://github.com/tdemin16/Continual-LayerNorm-Tuning}.}}
\end{abstract}

\section{Introduction}
\label{sec:intro}

The goal of class-incremental learning (IL)~\cite{rebuffi2017icarl} is to learn a classifier over a sequence of learning steps (or tasks), where each one contains a different set of classes. One of the main challenges of IL is to avoid forgetting old knowledge when learning new categories, a phenomenon called catastrophic forgetting~\cite{mccloskey1989catastrophic,kemker2018measuring}. One of the most effective ways to mitigate forgetting is to store a subset of old class data and use them to perform rehearsal of old knowledge~\cite{rebuffi2017icarl,prabhu2020gdumb,liu2020mnemonics,luo2023class,ghunaim2023real}. However, storing samples of old classes is not always possible due to potential memory constraints~\cite{fini2020online} or privacy issues~\cite{wang2022learning,zhu2021prototype}. 

Recent works~\cite{wang2022learning, wang2022dualprompt, smith2023coda} demonstrated how exploiting pre-trained Vision Transformers (ViT)~\cite{dosovitskiy2020image} can lead to 
competitive IL results even in a rehearsal-free setting. The key idea behind these models is to freeze the pre-trained backbone and learn specific input prompts for each learning step. During inference, the model selects which prompts to use based on step-specific keys and the activations of the pre-trained backbone for the current input. Notably, the visual prompts of one learning step do not interfere with the old ones~\cite{wang2022learning, wang2022dualprompt}, drastically reducing catastrophic forgetting. Despite their advantages, these methods are inefficient in terms of computational cost: the number of learnable parameters (e.g., prompts) impacts the results, and a large number of prompts is required to achieve state-of-the-art performance~\cite{smith2023coda}.

\begin{figure}
    \centering
    \includegraphics[width=1\linewidth]{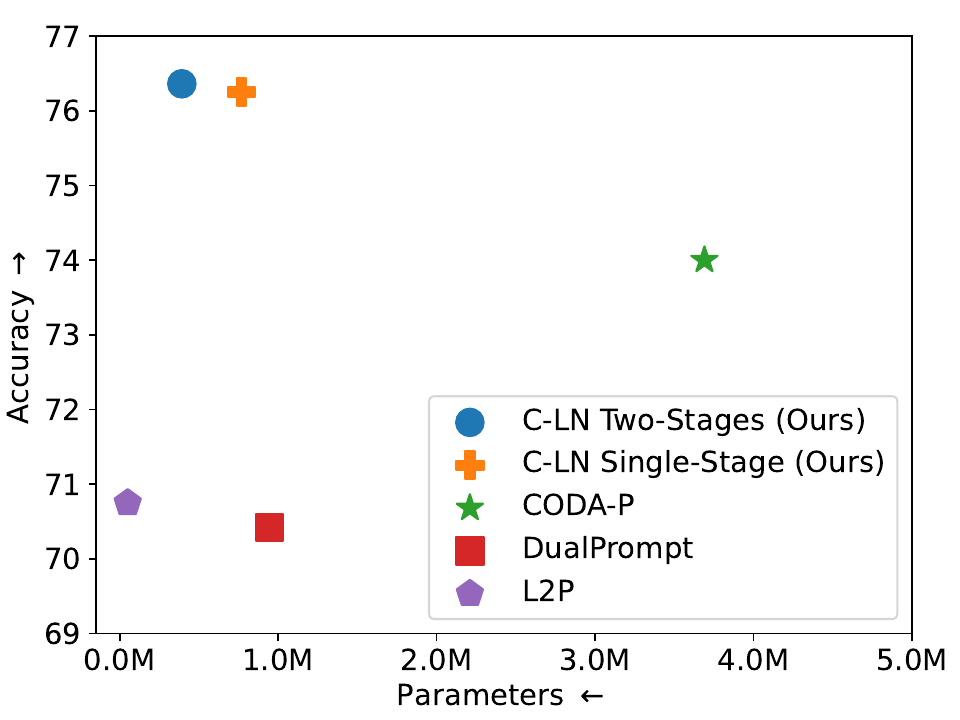}
    \caption{Continual-LayerNorm's efficiency compared to state-of-the-art methods. Our approach, C-LN, improves the accuracy in ImageNet-R over a 10-task benchmark while training fewer parameters.}
    \label{fig:params}
\end{figure}

In this work, we study whether it is possible to sidestep this computational burden while maintaining competitive performance. To achieve this, we propose to revisit and extend a simple transfer learning idea, i.e., learning specific normalization layers. 
Inspired by recent works on efficient ViT fine-tuning~\cite{basu2023strong}, we learn specific scale and bias parameters of LayerNorm \cite{ba2016layer} for each incremental learning step. As in prompt learning approaches~\cite{wang2022learning}, we 
learn task-specific keys, 
training them to mimic the output of the frozen pre-trained ViT for samples of their specific learning step. During inference, we then use the parameters whose associated key is the closest 
to the pre-trained model output for the given input sample. 
Finally, we make the classifier more robust to selection errors via a two-stage training pipeline, where we first use the ground-truth task identity to tune the 
LayerNorm parameters and then simulate the inference-time selection procedure to 
train the classifier. 
We name our approach \textbf{C}ontinual \textbf{L}ayer\textbf{N}orm Tuning, C-LayerNorm.

Experiments on CIFAR-100~\cite{krizhevsky2009learning} and ImageNet-R~\cite{hendrycks2021many}, 
show that C-LayerNorm 
achieves either the same or superior results of prompt tuning strategies, while providing up to 90\% reduction over the number of trained parameters 
compared to the state of the art, CODA-Prompt~\cite{smith2023coda}. This is reported in Fig.~\ref{fig:params} showing the accuracy vs the number of parameters trained for different approaches on the ImageNet-R dataset, with our C-LayerNorm achieving the best tradeoff. 

In summary, we make the following contributions:
\begin{itemize}
    \item We propose C-LayerNorm, a rehearsal-free continual learning method that tunes task-specific 
    LayerNorm parameters for each incremental learning step, selecting them during inference. 
    
    \item We design a variant of the approach for single-stage inference, where the selection is performed within each LayerNorm. Despite tuning more parameters, this strategy leads to 
    an inference speed two times faster than the current state of the art.

    \item Experiments on standard benchmarks show that both single- and two-stage approaches 
    consistently achieve state-of-the-art performance, 
    with the best 
    tradeoff between performance and the number of parameters.
\end{itemize}

\section{Related Work}
\label{sec:related}
\paragraph{Continual Learning.} 
Early approaches for continual learning tackled catastrophic forgetting through regularization objectives, e.g., preventing changes of model parameters important for old tasks \cite{kirkpatrick2017overcoming,zenke2017continual,chaudhry2018riemannian,aljundi2018memory} or via distilling the old model's activations \cite{li2017learning,rebuffi2017icarl,castro2018end,fini2022self,cermelli2020modeling,cermelli2021modeling}.
These methods can be coupled with a memory buffer, storing samples of old tasks/classes for performing knowledge rehearsal~\cite{prabhu2020gdumb,luo2023class,liu2020mnemonics,cha2021co2l,zhang2022simple}. 
While this strategy is extremely effective for incremental learning, both in terms of accuracy and efficiency \cite{ghunaim2023real,prabhu2023computationally}, it is not viable in case of privacy constraints, preventing storing data of past tasks. 
To deal with such cases, the community focused on rehearsal-free continual learning, with 
different works proposing various strategies, ranging from strong regularization objectives on the model's activations~\cite{fini2020online}, to self-supervised objectives~\cite{zhu2021prototype}, and modeling the semantic drift of the network's representation~\cite{yu2020semantic}. 
More recently, ViT-specific approaches showed that rehearsal is not needed to 
achieve state-of-the-art performance in continual learning. 
These models tune a subset of the network parameters for each task/incremental step, selecting the most relevant ones during inference.
The seminal work in this direction is Learning to Prompt~\cite{wang2022learning} which proposes to perform visual prompt tuning~\cite{jia2022visual} and prepend a set of learnable tokens as input to the Vision Transformer. 
Follow-up works improve either the way prompts are injected within the model~\cite{wang2022dualprompt} or trained across incremental learning steps~\cite{smith2023coda}. 
These approaches achieve impressive results but inherit a trade-off between several task-specific parameters and performance. 

Differently from these works, in this paper, we propose to learn specific normalization layers, showing that this strategy achieves either comparable or superior performance while requiring much fewer parameters than the state-of-the-art prompting methods~\cite{smith2023coda,wang2022dualprompt}.

\paragraph{Parameter-Efficient Fine-Tuning.} 
Due to its relevance for resource-constrained settings, parameter-efficient fine-tuning, i.e., tuning a pre-trained model on a specific task without either re-training it from scratch or adding a large number of parameters, is a widely studied topic in machine learning. In computer vision, early works explored the use of adapters to modify the activations of residual blocks in convolutional neural networks~\cite{rebuffi2017learning,rebuffi2018efficient}. Subsequent efforts further reduced the number of parameters, either by extracting task-specific subnetworks \cite{mallya2018piggyback,berriel2019budget,mancini2018adding,mancini2020boosting}  or by factorizing the network's layers~\cite{bulat2020incremental}. For vision transformers, multiple techniques have been borrowed from the natural language processing literature, such as standard~\cite{houlsby2019parameter} and low-rank adapters~\cite{hu2021lora}, or prompt-tuning via additional input tokens~\cite{li2021prefix,jia2022visual} or modifying the images themselves~\cite{bahng2022exploring}. 

In this work, we follow a different path, learning specific normalization parameters per incremental step. This shares similarities with previous works 
tuning specific batch-normalization parameters~\cite{bilen2017universal}, layer normalization ones~\cite{basu2023strong}, or by introducing covariance normalization layers~\cite{li2019efficient}. However, here we focus on a different problem formulation, demonstrating that normalization strategies are an effective way to perform rehearsal-free IL in vision transformers.

\section{Method}
\label{sec:prel}
In this section, we first describe the problem formulation (Sec.~\ref{sec:problem}), and how LayerNorm can be used for continual learning (Sec.~\ref{sec:LN-tune}). We then present our two-stage (Sec.~\ref{sec:2stage}) and single-stage~(Sec.~\ref{sec:1stage}) approaches. Finally, we describe our full training procedure and how we make the classifier more robust to selection errors during inference (Sec.~\ref{sec:training}).

\begin{figure*}
    \centering
    \begin{subfigure}[b]{0.49\textwidth}
        \centering
        \includegraphics[width=\textwidth]{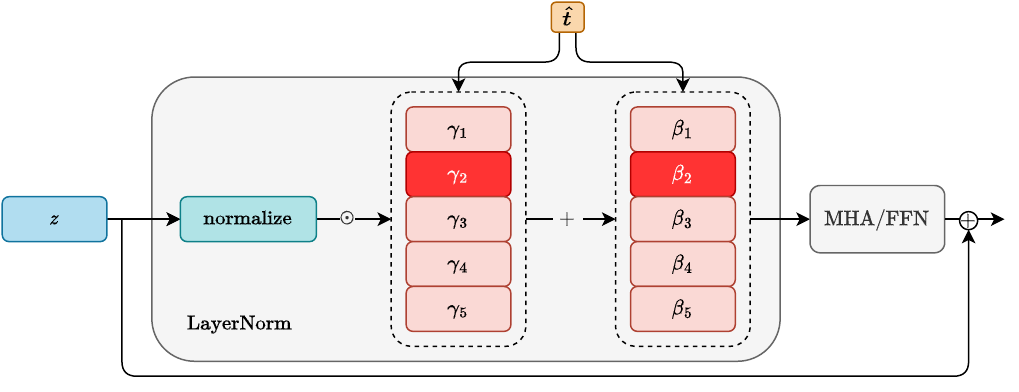}
        \caption{Two-Stages}
        \label{fig:method-two_stages}
    \end{subfigure}
    \hfill
    \begin{subfigure}[b]{0.49\textwidth}
        \centering
        \includegraphics[width=\textwidth]{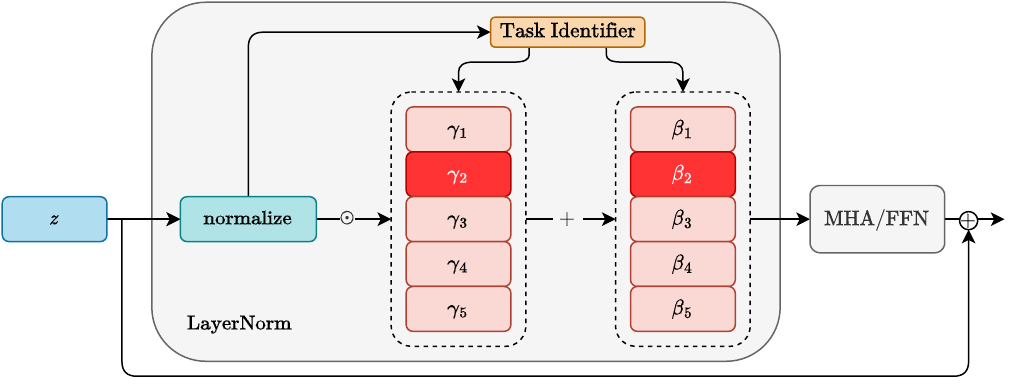}
        \caption{Single-Stage}
        \label{fig:method-single_stage}
    \end{subfigure}
    \caption{Two variants of our approach. On the left, every LayerNorm in the Two-Stages version must be provided with the task identity to select the set of parameters. On the right, instead, our Single-Stage LayerNorm estimates task identities on the fly. By doing so, we do not need to compute task identities a priori, thus almost halving the computational burden.}
    \label{fig:method}
\end{figure*}

\begin{figure}
    \centering
    \includegraphics[width=1\linewidth]{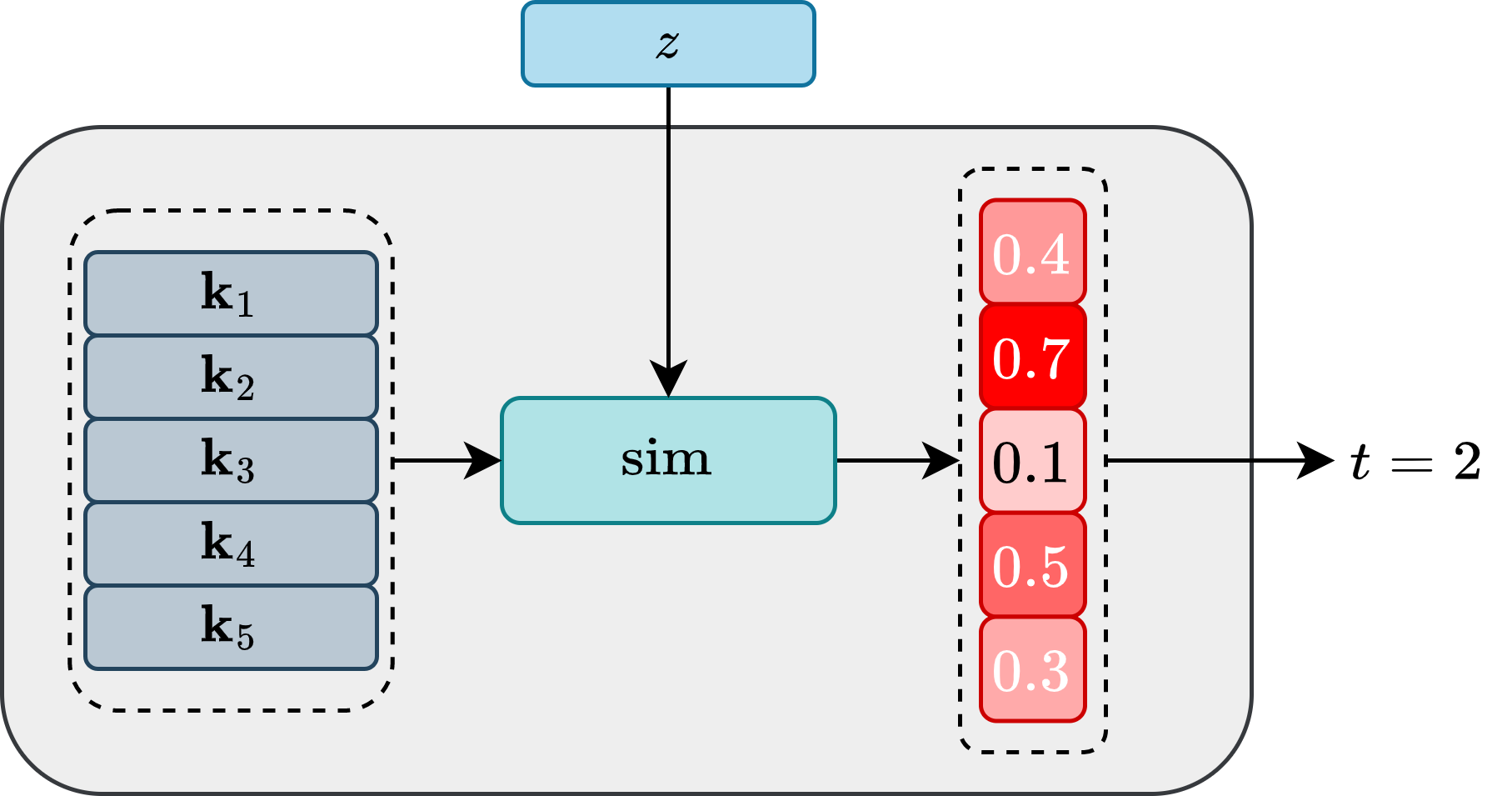}
    \caption{Task identifier structure. We compute the similarity between the feature vector $z$ and all the learned task vectors. Each comparison leads to a similarity score. We infer task identities as the index of the vector that has the highest similarity score with the input features.}
    \label{fig:identifier}
\end{figure}

\subsection{Notation and problem formulation}
\label{sec:problem}
The goal of continual learning is to learn a model from a sequence of non-stationary datasets, performing training over multiple steps.
Formally, we receive sequential data $\mathcal{D} = \{{D}_1,\cdots, {D}_T\}$, where $T$ is the number of incremental learning steps/tasks and ${D}_t$ is the dataset available at timestep $t$.
Each dataset is a collection of tuples such that ${D}_t = \{(\mathbf{x}^t_i, y^t_i)\}_{i=1}^{n_t}$.
$\mathbf{x}_i^t \in \mathcal{X}$ is the \textit{i}-th image of task $t$ in the image space $\mathcal{X}$, and $y_i^t\in \mathcal{Y}_t$ is its corresponding label in $\mathcal{Y}_t$ in the case of the image classification problem. Note that the semantic space of the dataset ${D}_t$, i.e., $\mathcal{Y}_t$, does not overlap with the one at a different learning step, i.e., $\mathcal{Y}_t\cap \mathcal{Y}_u = \varnothing$ for each step $t,u \in \{1,\cdots,T\}$ with $t\neq u$.

Given $\mathcal{D}$, we aim to learn a function $f^T_\theta: \mathcal{X} \rightarrow \mathcal{Y}$, parametrized by $\theta$ that maps images in the space $\mathcal{X}$ to labels in the set $\mathcal{Y}$, where $\mathcal{Y}=\bigcup_{t=1}^T \mathcal{Y}_t$. The literature distinguishes between
task- and class-incremental learning depending on the availability of task identities during inference. 
In this work, we focus on the latter, where task identities are unknown and $f^T_\theta$ performs predictions over the full set $\mathcal{Y}$. In the following, we will use the terms \textit{incremental step} and \textit{task} interchangeably. 

To tackle this problem, we split the parameters $\theta$ into two sets: $\theta_g$ containing global (frozen parameters) and a set of task-specific parameters $\theta_s$, fine-tuned for each learning step, i.e., $\theta_s = \{\theta_s^1,\cdots, \theta_s^T\}$. 
While standard approaches use prompts as $\theta_s$, their performance is influenced by the number of task-specific parameters they train~\cite{smith2023coda}.
In this work, we take a different direction, 
and show that tuning LayerNorm can reduce the number of learnable parameters without sacrificing the final accuracy.

\subsection{Continual Learning with LayerNorm}
\label{sec:LN-tune}
Inspired by the multi-domain learning~\cite{bilen2017universal} and the transfer learning~\cite{basu2023strong} literatures, we address catastrophic forgetting by learning specific normalization layer weights for each incremental learning step.
In the case of ViTs, the normalization layers are LayerNorm ones~\cite{ba2016layer}, in which we tune the scale and bias parameters. 
Formally, given as input a feature vector $z$, LayerNorm~\cite{ba2016layer} is defined as:
\begin{equation}
    \label{eq:LN}
    \text{LayerNorm}(z) = \frac{z - \text{E}[z]}{\sqrt{\text{Var}[z] + \epsilon}} \odot \gamma^\ell + \beta^\ell,
\end{equation}
where $\gamma^\ell$ and $\beta^\ell$ are the learnable scale and bias parameters for layer $\ell$, and $\epsilon$ is a small value to prevent numerical problems. For the purpose of continual learning, we aim to learn a specific LayerNorm for each incremental learning step. We thus revise Eq.~\eqref{eq:LN} as
\begin{equation}
    \label{eq:C-LN}
    \text{C-LayerNorm}(z,t) = \frac{z - \text{E}[z]}{\sqrt{\text{Var}[z] + \epsilon}} \odot \gamma_t^\ell + \beta_t^\ell,
\end{equation}
where $t$ is the incremental step indicator and $\gamma_t^\ell$ and $\beta_t^\ell$ are its specific scale and bias parameters for layer $\ell$. 
{Since we allocate a set of LayerNorm parameters for each task, during inference we need to infer task identities
to select the best weights for the given input sample.}
In the following, we describe two ways of selecting parameters during inference: two-stage (Fig.~\ref{fig:method-two_stages}), where we first infer the identity and then perform the prediction, and single-stage (Fig.~\ref{fig:method-single_stage}), where we estimate the identity within each C-LayerNorm.

\subsection{Two-stage C-LayerNorm}
\label{sec:2stage}
In this case, we follow prompt-learning approaches for continual learning~\cite{wang2022learning} and instantiate a learnable key vector for each incremental step (Fig.~\ref{fig:identifier}). The key vector will then be used during inference to estimate the task identity. Formally, for each dataset $D_t$, we instantiate a key parameter $\theta^t_{\mathtt{SEL}}\in\mathbb{R}^d$ where $d$ is the output dimension of the $\mathtt{cls}$ token of the frozen pre-trained model. We learn $\theta^t_{\mathtt{SEL}}$ by minimizing the following objective:
\begin{equation}
    \label{eq:loss-2s}
    \mathcal{L}^{\text{2-stage}}_{\mathtt{key}}(\theta^t_\mathtt{SEL}) = \frac{1}{|D_t|} \sum_{(x,y)\in D_t} 1-\frac{f^0_\theta(x)^\top\,\theta^t_\mathtt{SEL}}{||f^0_\theta(x)||\cdot||\theta^t_{\mathtt{SEL}}||},
\end{equation}
where, for simplicity, $f_\theta^0$ denotes the pre-trained model, mapping images to their corresponding $\mathtt{cls}$ embeddings. Eq.~\eqref{eq:loss-2s} maximizes the average cosine similarity between the task-specific key $\theta^t_{\mathtt{SEL}}$ and the embedding of the dataset samples, as extracted from the pre-trained model. 

During inference, we estimate the task identity $\hat{t}$ (and thus the specific set of C-LayerNorm parameters to use) by computing the cosine similarity between the keys and the embedding of the input sample $\mathbf{x}$, i.e.,
\begin{equation}
    \label{eq:key-2s}
    \hat{t} = \underset{t\in\{1,\cdots,T\}}{\arg\max} \frac{f^0_\theta(\mathbf{x})^\top\,\theta^t_\mathtt{SEL}}{||f^0_\theta(\mathbf{x})||\cdot||\theta^t_\mathtt{SEL}||}.
\end{equation}
Note that, given a sample, all C-LayerNorm of the model uses the same unique task identity, thus we first estimate the task ID with a forward pass over $f_\theta^0$ and then perform a second forward on $f_\theta^T$ to get the semantic prediction (Fig.~\ref{fig:method-two_stages}).

\subsection{Single-stage C-LayerNorm}
\label{sec:1stage}
An alternative to the two-stage procedure is to compute the task ID on the fly, on each C-LayerNorm, avoiding multiple sequential forward passes (Fig.~\ref{fig:method-single_stage}). 
To achieve this, we instantiate one task-specific key for each C-LayerNorm, performing a similar procedure as in Section~\ref{sec:2stage}, but independently at each layer. 
Formally, let us denote as $z_{\ell}$ the activation in input to the C-LayerNorm $\ell$, and as $\theta^{t,\ell}_\mathtt{SEL}$ the selection key at layer $\ell$ for step $t$. 
Similarly to the previous case, we perform the parameters' selection at layer $\ell$ as
\begin{equation}
    \label{eq:key-1sL}
    \hat{t}_\ell = \underset{t\in\{1,\cdots,T\}}{\arg\max} \frac{(\alpha_\ell^t\odot {z^0_{\ell}})^\top\,\theta^{t,\ell}_\mathtt{SEL}}{||\alpha_\ell^t\odot z^0_{\ell}||\cdot||\theta^{t,\ell}_\mathtt{SEL}||}.
\end{equation}
where $z^0_{\ell}$ is the $\mathtt{cls}$ token embeddings at the same layer as given by the pre-trained model $f^0_\theta$, and $\alpha_\ell^t\in \mathbb{R}^d$ is a learnable attention vector scaling the values of the input. Similarly to \cite{smith2023coda}, we found it beneficial to introduce this attention vector,  adding flexibility to the selection procedure.

We train both the keys and the attentions using an input dataset $D_t$, maximizing the cosine similarity between the scaled $\mathtt{cls}$ token embeddings and the keys:
\begin{equation}
    \label{eq:loss-1s}
    \mathcal{L}^{\text{1-stage}}_{\mathtt{key},\ell}(\theta^{t,\ell}_\mathtt{SEL},\alpha^{t}_\ell) = \frac{1}{|D_t|} \sum_{(x,y)\in D_t} 1-\frac{({\alpha_\ell^t\odot z^0_\ell})^\top\,\theta^{t,\ell}_\mathtt{SEL}}{||\alpha_\ell^t\odot z^0_\ell||\cdot||\theta^{t,\ell}_{\mathtt{SEL}}||}
\end{equation}
and thus, considering $L$ C-LayerNorm layers:
\begin{equation}
    \label{eq:loss-1s-layers}
    \mathcal{L}^{\text{1-stage}}_{\mathtt{key}}(\theta^{t}_\mathtt{SEL}) = \frac{1}{L} \sum_{\ell= 1}^{L} \mathcal{L}^{\text{1-stage}}_{\mathtt{key},\ell}(\theta^{t,\ell}_\mathtt{SEL},\alpha^{t}_\ell),
\end{equation}
where here, for simplicity, we denote as $\theta^{t}_\mathtt{SEL}$ the full set of task-specific selection parameters, i.e., $\theta^{t}_\mathtt{SEL}=\{\theta^{t,\ell}_\mathtt{SEL}, \alpha^{t}_\ell\}_{\ell=1}^L$. 
While this formulation requires additional parameters, 
it has two main advantages: i) it almost halves the inference time, as there is no need to perform sequential forward passes; ii) it allows training all parameters end-to-end, as the gradient flows to the selection keys and attention vectors through the intermediate representations.

\subsection{Training pipeline}
\label{sec:training}
Summarizing the steps before, given a dataset ${D}_t$ at timestep $t$, we define its specific set of parameters as $\theta_s^t = \{\theta_\mathtt{LN}^t, \theta_\mathtt{SEL}^t,\theta_\mathtt{CLS}^t\}$, where $\theta_\mathtt{LN}^t$ are the parameters of C-LayerNorm for all $L$ layers of the network, $\theta_\mathtt{LN}^t = \{\gamma^\ell_t,\beta^\ell_t\}_{\ell=1}^L$, $\theta_\mathtt{SEL}^t$ are the parameters involved in the selection process (i.e. keys for the two-stage variant, layer-wise keys, and attention vectors for the single stage one).
Moreover, at each time step, we introduce $\theta_\mathtt{CLS}^t$, the parameters of the classifier necessary to recognize the new classes. 

Practically, we may define a single objective, learning $\theta_\mathtt{CLS}^t$ and $\theta_\mathtt{LN}^t$ together. However, we found that the mismatch between the set of parameters used during training (i.e., ground-truth task) and inference (i.e., estimated), results in generalization problems, harming the performance of the model. This is shown in Figure~\ref{fig:barplot}, where there is a clear gap between the performance of the oracle task selector and the estimated identities via task-specific keys.  To reduce this issue we propose 
a two-stage training procedure, where we first tune the task-specific parameters and then fine-tune the classifier using the same parameters selection procedure adopted during inference.

\paragraph{C-LayerNorm parameters training.} To tune the task-specific parameters, we use ${D}_t$ and minimize
\begin{equation}
    \label{eq:full-objective}
   \mathcal{L}(\theta_s^t) = \mathcal{L}_\mathtt{CE}(\theta^t_\mathtt{LN}, \theta^t_{\mathtt{CLS}}) + \mathcal{L}_\mathtt{key}(\theta^t_\mathtt{SEL}),
\end{equation}
where $\mathcal{L}_\mathtt{CE}$ is the standard cross-entropy loss:
\begin{equation}
    \label{eq:ce-loss}
   \mathcal{L}_{\text{CE}}(\theta^t_\mathtt{LN}, \theta^t_{\mathtt{CLS}}) = - \frac{1}{|{D}_t|} \sum_{(x,y)\in{D}_t} \log f^t_\theta(x)[y]. 
\end{equation}
Above, $f^t_\theta(x)[y] $ is the prediction of the model for label $y$. At each incremental step, we minimize Eq.~\eqref{eq:full-objective} only w.r.t. the task-specific parameters in $\theta$, i.e., $\theta_s^t$. Following \cite{smith2023coda}, also for the classifier we update only the parameters of new classes $\theta^t_{\mathtt{CLS}}$, excluding the others from the backpropagation step. Note that, during the forward pass of C-LayerNorm, we use the ground-truth task identity to ensure that these parameters are optimal and truly specific for the current task.

\paragraph{Classifier refinement.} After estimating the task-specific parameters for task selection and C-LayerNorm, we re-initialize the classifier and fine-tune it to be more robust to selection errors during inference. Specifically, we use the cross-entropy loss as defined in Eq.~\eqref{eq:ce-loss}, training only ${\theta}_\mathtt{CLS}^t$ with all the other parameters frozen (i.e., $\theta_{\mathtt{LN}}^t$ and $\theta_{\mathtt{SEL}}^t$) while performing task-selection as during inference time. Note that, by not using the ground-truth identifier, we are exposing the classifier to the potential selection mistakes that the model will experience during inference. This makes the model more robust to selection errors, providing improvements in the overall performance, as shown in Figure~\ref{fig:barplot} and in Sec.~\ref{sec:exp}.

\section{Experiments}
\label{sec:exp}
Here, we first describe our experimental setting (Sec.~\ref{sec:exp-details}) and compare C-LayerNorm with the state-of-the-art in IL (Sec.~\ref{sec:exp-sota}). Finally, we analyze how the classifier refinement influences the performance, 
and, qualitatively, the learned task-specific parameters (Sec.~\ref{sec:exp-ablations}).

\subsection{Experimental setting}
\label{sec:exp-details}

\paragraph{Datasets.} We conduct experiments on two benchmark datasets used by previous works~\cite{wang2022learning,wang2022dualprompt,smith2023coda}.
The first is \textbf{CIFAR-100}~\cite{krizhevsky2009learning}, a 60$k$ samples dataset that spans 100 classes with 32$\times32$ images.
As the images have a smaller dimension compared to the input size of ViT, we upscale them to the expected input size.
The second dataset we use is \textbf{ImageNet-R}~\cite{hendrycks2021many}.
This dataset is challenging, as it contains 30$k$ images spanning 200 classes and 15 different visual domains.
This latter characteristic allows for testing the model in a scenario where the pre-training dataset distribution (i.e., ImageNet-1k~\cite{deng2009imagenet}) differs from the target one.

\paragraph{Metrics.} We use the same evaluation metrics of previous works in this field, namely
average accuracy and forgetting.
The first is defined as the average accuracy on the last trained model, evaluated on the union of all classes learned across all incremental steps.
On the other hand, forgetting measures the average amount of information loss.
We refer to \textit{Wang} \etal~\cite{wang2022dualprompt} for further details.

\paragraph{Baselines.} 
We consider various baselines for our experiments. The main 
competitors are state-of-the-art approaches for rehearsal-free continual learning, namely Learning to Prompt (L2P)~\cite{wang2022learning}, DualPrompt~\cite{wang2022dualprompt}, and CODA-Prompt~\cite{smith2023coda}.
L2P allocates distinct prompts to different tasks, inferring task identities using selection parameters.
DualPrompt~\cite{wang2022dualprompt} expands L2P by replacing 
prompt-tuning~\cite{lester2021power} with
prefix-tuning~\cite{li2021prefix}.
CODA-Prompt~\cite{smith2023coda} trains the selection parameters end-to-end, with dynamic prompt allocation. For fairness and compatibility of the results\footnote{There is a mismatch in the computation of the forgetting metric between \cite{smith2023coda} and \cite{wang2022dualprompt}. As in \cite{wang2022dualprompt}, we follow the original definition in \cite{chaudhry2018riemannian}.}, we re-ran these methods using their original code. 

To provide a reference upper bound, we report the results of training a model on the whole dataset at once, either 
fine-tuning all the parameters in the ViT (UB), or just LayerNorm ones (LN-Tune~\cite{basu2023strong}).
Similarly, as lower-bound, we consider fine-tuning the ViT on each incremental learning step, without strategies to prevent forgetting (FT).
Following previous works, we also report the results of five rehearsal-based approaches, ER~\cite{chaudhry2019tiny}, BiC~\cite{wu2019large}, GDumb~\cite{prabhu2020gdumb}, DER~\cite{buzzega2020dark}, and Co$^2$L~\cite{cha2021co2l}, evaluating them with a replay buffer of 1000 and 5000 images.

\paragraph{Implementation details.} We follow previous work~\cite{smith2023coda,wang2022learning}, and perform our experiments using a ViT-Base backbone~\cite{dosovitskiy2020image} pre-trained on ImageNet-1k~\cite{deng2009imagenet}. To train our method, we use the Adam optimizer~\cite{kingma2014adam} with standard beta values, $\beta_1=0.9$ and $\beta_2=0.999$.
The batch size is set to 128 images for all experiments, and all the images are resized to 224$\times$224, training the models for 5 epochs on each step.
The learning rate is set to $1e^{-2}$ for ImageNet-R~\cite{hendrycks2021many} and $5e^{-3}$ for CIFAR-100~\cite{krizhevsky2009learning}.
 For both single- and two-stage variants, the classifier is trained with a
 constant learning rate of $1e^{-3}$ for both datasets.
Note that, compared to prompt-tuning methods
~\cite{wang2022learning,smith2023coda,wang2022dualprompt}, our approach does not introduce any additional hyperparameter.

\subsection{Comparison with the state of the art}
\label{sec:exp-sota}
\paragraph{10-task benchmarks.} In Table~\ref{tab:results}, we report the results of our approach, C-LayerNorm (C-LN), and all baselines for CIFAR-100 and ImageNet-R on a 10-task benchmark (i.e., splitting the classes into 10 groups and learning one group after the other).
From top to bottom, we show the upper bound, rehearsal-based methods,
regularization-based ones, prompt-based models, and C-LayerNorm (C-LN).
For each method and setting, we report accuracy and forgetting\footnote{GDumb~\cite{prabhu2020gdumb}, has
no forgetting value as it trains the network from scratch after each task.}.

\begin{table*}
    \centering
    \caption{Results on CIFAR-100 and ImageNet-R 10-task benchmarks in class-incremental learning. Both variants of our method perform on par with the current state-of-the-art in the first dataset, while setting a new SOTA in ImageNet-R.}
    \begin{tabular}{c c c c c c c}
    \hline
        \multirow{2}{*}{\textbf{Method}} & \multirow{2}{*}{\textbf{Buffer size}} & \multicolumn{2}{c}{\textbf{CIFAR-100}} & \multirow{2}{*}{\textbf{Buffer size}} & \multicolumn{2}{c}{\textbf{ImageNet-R}}\\
        & & Accuracy $\uparrow$ & Forgetting $\downarrow$ & & Accuracy $\uparrow$ & Forgetting $\downarrow$ \\
    \hline
        UB & - & $90.85$\scriptsize$\pm0.12$ & - & - & $79.13$\scriptsize$\pm0.18$ & - \\
        LN-Tune & - & 92.05 & - & - & 81.18 & - \\
    \hline
        ER & \multirow{5}{*}{1000} & $67.87$\scriptsize$\pm0.57$ & $33.33$\scriptsize$\pm1.28$  & \multirow{5}{*}{1000} & $53.13$\scriptsize$\pm1.29$ & $35.38$\scriptsize$\pm0.52$ \\
        BiC & & $66.11$\scriptsize$\pm1.76$ & $35.24$\scriptsize$\pm1.64$ & & $52.14$\scriptsize$\pm1.08$ & $36.70$\scriptsize$\pm1.05$\\
        GDumb & & $67.14$\scriptsize$\pm0.37$ & -  & & $38.32$\scriptsize$\pm0.55$ & - \\
        DER & & $61.06$\scriptsize$\pm0.87$ & $39.87$\scriptsize$\pm0.99$ & & $55.47$\scriptsize$\pm1.31$ & $34.64$\scriptsize$\pm1.50$ \\
        Co$^2$L & & $72.15$\scriptsize$\pm1.32$ & $28.55$\scriptsize$\pm1.56$ & & $53.45$\scriptsize$\pm1.55$ & $37.30$\scriptsize$\pm1.81$ \\
    \hline
        ER & \multirow{5}{*}{5000} & $82.53$\scriptsize$\pm0.17$ & $16.46$\scriptsize$\pm0.25$ & \multirow{5}{*}{5000} & $65.18$\scriptsize$\pm0.40$ & $23.31$\scriptsize$\pm0.89$ \\
        BiC & & $81.42$\scriptsize$\pm0.85$ & $17.31$\scriptsize$\pm1.02$ &  & $64.63$\scriptsize$\pm1.27$ & $22.25$\scriptsize$\pm1.73$\\
        GDumb & & $81.67$\scriptsize$\pm0.02$ & - &  & $65.90$\scriptsize$\pm0.28$ & - \\
        DER & & $83.94$\scriptsize$\pm0.34$ & $14.55$\scriptsize$\pm0.73$ &  & $66.73$\scriptsize$\pm0.87$ & $20.67$\scriptsize$\pm1.24$ \\
        Co$^2$L & & $82.49$\scriptsize$\pm0.89$ & $17.48$\scriptsize$\pm1.80$ &  & $65.90$\scriptsize$\pm0.14$ & $23.36$\scriptsize$\pm0.71$ \\
    \hline
        FT & \multirow{7}{*}{0} & $33.61$\scriptsize$\pm0.85$ & $86.87$\scriptsize$\pm0.20$  & \multirow{7}{*}{0} & $28.87$\scriptsize$\pm1.36$ & $63.80$\scriptsize$\pm1.50$ \\
        EWC & & $47.01$\scriptsize$\pm0.29$ & $33.27$\scriptsize$\pm1.17$ &  & $35.00$\scriptsize$\pm0.43$ & $56.16$\scriptsize$\pm0.88$\\
        LwF & & $60.69$\scriptsize$\pm0.63$ & $27.77$\scriptsize$\pm2.17$  &  & $38.54$\scriptsize$\pm1.23$ & $52.37$\scriptsize$\pm0.64$ \\
        L2P & & $83.60$\scriptsize$\pm0.59$ & \underline{6.06}\scriptsize$\pm0.18$  &  & $70.75$\scriptsize$\pm0.23$ & $\textbf{4.53}$\scriptsize$\pm0.64$ \\
        DualPrompt & & $81.34$\scriptsize$\pm0.24$ & $7.81$\scriptsize$\pm1.38$  &  & $70.42$\scriptsize$\pm{0.21}$ & $5.86$\scriptsize$\pm0.74$ \\
        CODA-P & & \underline{86.84}\scriptsize$\pm0.41$ & \textbf{5.76}\scriptsize$\pm0.64$  & & $74.00$\scriptsize$\pm0.42$ & $6.62$\scriptsize$\pm0.57$ \\
    \hline
        \textbf{C-LN Two-Stages} & \multirow{2}{*}{0} & $\textbf{86.95}$\scriptsize$\pm0.37$ & $6.98$\scriptsize$\pm0.43$ & \multirow{2}{*}{0} & $\textbf{76.36}$\scriptsize$\pm0.51$ & $8.31$\scriptsize$\pm1.28$\\
        \textbf{C-LN Single-Stage} & & $86.06$\scriptsize$\pm0.83$ & $8.60$\scriptsize$\pm1.68$ &  & $\textbf{76.25}$\scriptsize$\pm0.79$ & $9.01$\scriptsize$\pm1.08$ \\
    \hline
    \end{tabular}
    \label{tab:results}
    \vspace{-0.25cm}
\end{table*}

Overall, C-LN achieves the best accuracy (or comparable) across all settings, for both the two-stage and single-stage variants. There is a large gap between the performance of rehearsal-based approaches and our model, with the best-performing competitor of this category (DER) achieving accuracy 3\% lower on CIFAR-100 (83.94\% vs. 86.95\%) and almost 10\% on ImageNet-R (66.73\% vs. 76.36\%) w.r.t. our two-stage variant, despite using a memory buffer of 5000 samples. Comparisons are striking for a reduced buffer of 1000 samples, with C-LN outperforming DER by almost $30\%$ on ImageNet-R.
We remark that our method does not store any old task data, being more viable in applications with privacy or memory constraints.

Comparing our approach with other rehearsal-free methods, C-LN still achieves the best accuracy, with a 2\% improvement over CODA-Prompt and almost 6\% over DualPrompt and L2P on the challenging ImageNet-R. On CIFAR-100, the gap between the models is reduced, with two-stage C-LN achieving the same accuracy as CODA-Prompt while still outperforming L2P (+3.3\%) and DualPrompt (+5.6\%). These results are remarkable, as the approaches share the same principles (e.g., selecting task-specific parameters) but they differ in the parameters tuned (prompts vs LayerNorm). Our results demonstrate the benefit of simple LayerNorm tuning vs. more complex strategies. 
It is also worth highlighting that C-LN single-stage and two-stage achieve comparable results, demonstrating that we can tradeoff the number of parameters for faster inference time without losing on the final performance.

While our model achieves good results w.r.t. accuracy, it shows higher forgetting than the prompt-learning strategies, which might be a consequence of the stricter parameter isolation provided by the latter. Nevertheless, the higher accuracy shows that our model achieves a better stability/plasticity tradeoff, thus equally balancing the preservation of old knowledge and the learning of new concepts.

\vspace{-0.5cm}
\paragraph{5-20 tasks on ImageNet-R.} To further test both variants of our method, we performed experiments on ImageNet-R varying the number of incremental learning steps, reducing them to 5 (40 classes per task) and increasing them to 20 (10 classes per task), following~\cite{smith2023coda}. The results are shown in
Table~\ref{tab:tasks} in terms of accuracy, for both our model and the
prompt-based baselines. Even in these settings, our model achieves the best results, with the two-stage variant outperforming the previous state-of-the-art (CODA-Prompt) by almost 4\% with 5 steps and 1\% with 20. The same applies to our single-stage variant, outperforming CODA-Prompt by more than 3\% with 5 tasks and almost 1\% with 20. The gap is even more evident with L2P and DualPrompt, achieving results that are 7\% (6\% for the single-stage) lower than ours on 5 tasks, and 3\% lower for the 20 tasks case. These results confirm that the effectiveness of our model and the benefits of learning task-specific LayerNorm generalizes across different scenarios, and it is not affected by the number of incremental learning steps or classes per step.

\begin{table}
    \centering
    \caption{Comparison on ImageNet-R by changing the number of tasks. The accuracy of our method is higher on average than the baseline methods.}
    \begin{tabular}{ c c c c }
    \hline
        \multirow{2}{*}{\textbf{Method}} & \multicolumn{3}{c}{\textbf{Number of tasks}} \\
        & 5 $\uparrow$ & 10 $\uparrow$ & 20 $\uparrow$ \\
    \hline
        UB & 79.13 & 79.13 & 79.13 \\
        LN-Tune & 81.18 & 81.18 & 81.18 \\ 
    \hline
        L2P& $72.12$\scriptsize$\pm{0.44}$ & $70.75$\scriptsize$\pm{0.23}$ & $67.12$\scriptsize$\pm{0.36}$ \\
        DualPrompt & $71.83$\scriptsize$\pm{0.40}$ & $70.42$\scriptsize$\pm{0.21}$ & $68.15$\scriptsize$\pm{0.43}$ \\
        CODA-P & $75.25$\scriptsize$\pm{0.15}$ & $74.00$\scriptsize$\pm{0.32}$ & $70.74$\scriptsize$\pm{0.41}$ \\
    \hline
        \textbf{C-LN Two-S} & $\textbf{79.21}$\scriptsize$\pm{0.84}$ & \textbf{76.36}\scriptsize$\pm{0.51}$ & \textbf{71.72}\scriptsize$\pm{0.47}$ \\
        \textbf{C-LN Single-S} & \underline{78.43}\scriptsize$\pm{0.67}$ & \textbf{76.25}\scriptsize$\pm{0.79}$ & \textbf{71.62}\scriptsize$\pm{0.38}$ \\
    \hline
    \end{tabular}
    \label{tab:tasks}
\end{table}

\paragraph{Computational cost.} As analyzed in previous work~\cite{smith2023coda}, the number of learnable task-specific parameters plays a key role in the performance of rehearsal-free methods. Here we investigate whether the effectiveness of our model is linked to this aspect, reporting accuracy vs the number of trainable parameters in Table~\ref{tab:params}, for the ImageNet-R 10 tasks setting. As the table shows, our best-performing variant, i.e., the two-stage variant has a much lower number of learnable parameters than the previous state-of-the-art, CODA-Prompt (0.39M vs. 3.69M), despite achieving either comparable or superior results. While L2P uses much fewer parameters than our method (i.e., 0.05M), its performance has a large gap with ours (i.e., -5.6\% accuracy). Notably, increasing the number of parameters alone does not ensure an increase in performance, with DualPrompt using more parameters than our approach but achieving results comparable to L2P. These results confirm the benefit of using LayerNorm in rehearsal-free continual learning, achieving the best trade-off between performance and number of parameters. 

While all approaches discussed so far perform predictions via a two-stage pipeline, our single-stage variant provides prediction in a single forward pass. This results in a speed-up during inference, with our single-stage variant being 2$\times$ faster than our two-stage one. This speed is reflected also during training, where we witnessed a 1.6$\times$ speed-up compared to our two-stage variant. However, this advantage comes with a cost in terms of memory, almost doubling its requirement. Despite this, the single-stage variant still requires fewer parameters than DualPrompt and CODA-Prompt, while achieving either superior or comparable accuracy. Practically, one may pick one of the two C-LayerNorm variants based on the memory and/or speed required in the particular application scenario.

As a final note, it is worth highlighting that for the 5-task setting in Table~\ref{tab:tasks}, the performance of the Single-stage variant is lower than the two-stage one (i.e., 78.43\% accuracy vs. 79.21\%). This is surprising, as the single-stage variant has more flexibility in parameter selection since it might pick a different task ID at each layer. At the same time, we conjecture that this flexibility may cause an inherent distribution shift when multiple tasks are selected in subsequent layers, something that the classifier refinement step is not able to fix. While it is not trivial (i.e., as different layers focus on different features) encouraging a consistent task ID selection may further improve the performance of the single-stage variant.

\begin{table}
    \centering
    \caption{Accuracy vs. number of parameters in ImageNet-R 10-task benchmark. Our approach obtains a new state-of-the-art while reducing the number of parameters trained.}
    \begin{tabular}{c c c}
        \hline
        \textbf{Method} & \textbf{Accuracy} $\uparrow$ & \textbf{Param trained} $\downarrow$ \\
        \hline
        L2P & $70.75$\scriptsize$\pm{0.23}$ & $0.05M$ \\
        DualPrompt & $70.42$\scriptsize$\pm{0.21}$ & $0.94M$ \\
        CODA-P & $74.00$\scriptsize$\pm0.32$ & $3.69M$ \\
        \hline
        \textbf{C-LN Two-S} & $\textbf{76.36}$\scriptsize$\pm0.51$ & $\textbf{0.39M}$ \\
        \textbf{C-LN Single-S} & $\textbf{76.25}$\scriptsize$\pm0.79$ & $0.77M$ \\
        \hline
    \end{tabular}
    \label{tab:params}
\end{table}

\subsection{Analyses of C-LayerNorm}
\label{sec:exp-ablations}

\begin{figure}
    \centering
    \includegraphics[width=1\linewidth]{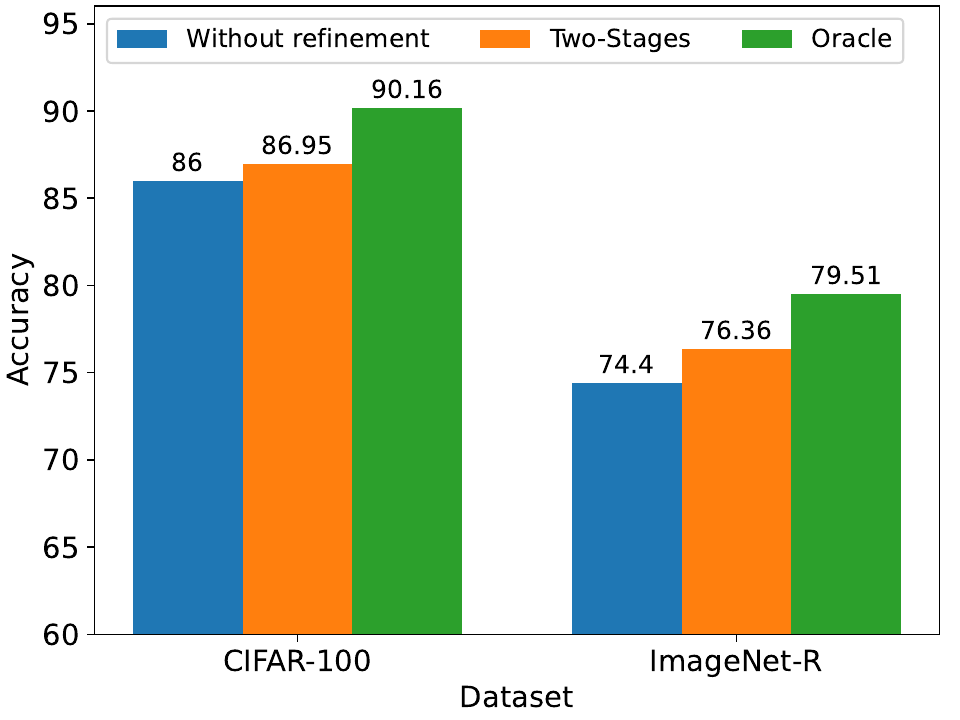}
    \caption{Classifier refinement effect. By training the classifier using inferred task identities instead of the ground truth ones, we improve the performance for both CIFAR-100 and ImageNet-R.}
    \label{fig:barplot}
\end{figure}

\paragraph{Impact of the classifier refinement step.}
As described in  Sec.~\ref{sec:training}, one of the key elements of our approach is the 
 classifier refinement step. In this step, we train only the classifier while selecting the C-LN parameters with the same procedure adopted during inference. The idea of this step is to make the classifier more robust to potential mistakes in the selection of the parameters during inference. 
 We analyze the impact of this choice in Figure~\ref{fig:barplot}, for the two-stage variant, showing the results on the 10-task settings without refinement (blue), with refinement (orange), and an oracle using the ground-truth task-specific parameters (green). 

Comparing the performance of our model without refinement and the oracle, we can see how correct parameter selection matters. Despite the two models differing only on the C-LayerNorm's parameter, the accuracy of the oracle is 4\% higher on CIFAR-100 (90.16\% vs. 86\%) and 5\% higher on ImageNet-R (79.51\% vs. 74.4\%) than the accuracy of our model without classifier refinement. 
These gaps suggest how mistakes in the parameters selection have a major impact on the final results, and should be taken into account in the model design. 

In this context, our classifier refinement step partly addresses this issue, with an improvement of almost 2\% on ImageNet-R and of 1\% on CIFAR-100 over the base model not adopting it. These results are promising, as they show that, by exposing the classifier during training to potential selection errors, it becomes more robust. At the same time, there is still a large gap between our full model and the oracle, making this a promising direction for future work.

\begin{figure}
    \centering
    \includegraphics[width=1.\linewidth]{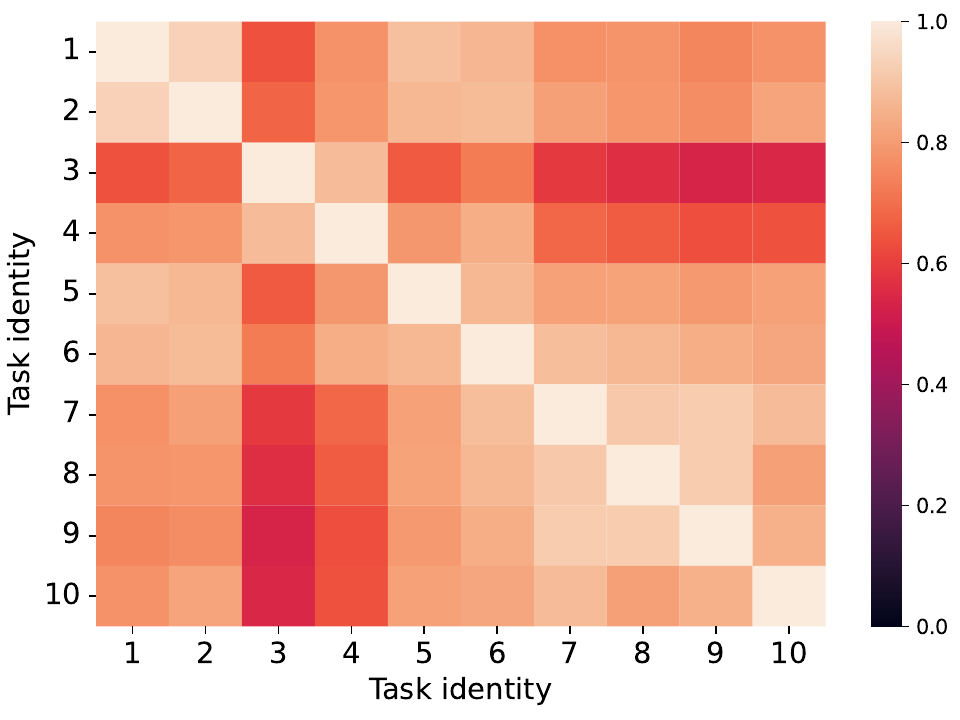}
    \caption{Similarity matrix of selection parameters on ImageNet-R 10-task benchmark. The learned selection vectors have a high inter-task similarity. That results in high pairwise similarity scores between the learned vectors.}
    \label{fig:heatmap}
\end{figure}

\begin{figure}
    \centering
    \includegraphics[width=1\linewidth]{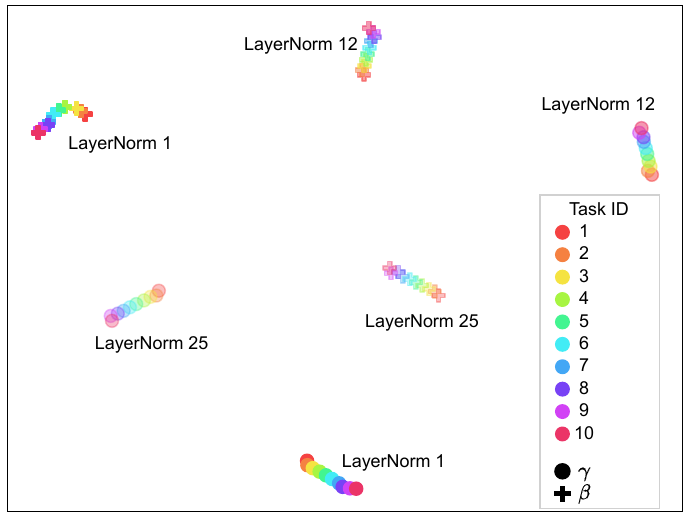}
    \caption{t-SNE on the learned weights and biases in the \nth{1}, \nth{12}, and \nth{25} LayerNorm layers in our approach. Representations have a high inter-task similarity. Such a property is desirable since the Task Identifier introduces errors in the weight selection. Such similar representations attenuate the error's magnitude, thus achieving consistent performances.}
    \label{fig:tsne}
\end{figure}

\paragraph{Learned task-specific parameters.} One assumption behind using task-specific parameters for continual learning, is that each set of parameters captures the peculiarities of the dataset it is exposed to. Here, we test this assumption, providing qualitative analyses of the parameters' similarity at different layers. The results are shown in Figure~\ref{fig:tsne}, where we report a t-SNE~\cite{van2008visualizing} of the scale and bias parameters of LayerNorm at different ViT layers (initial, middle, end, i.e., 1, 12, 25), for our two-stage variant and the 10-task {ImageNet-R} benchmark. As the figure shows, there is a pattern in the similarity of the parameters across tasks, with neighboring tasks (i.e., learned sequentially) being generally close in terms of learned parameters. This applies to both the scale and bias of LayerNorm. This behavior hints that these tasks need similar feature extractors and that subsequent tasks are more similar than randomly picked pairs. 

We verify this in Figure~\ref{fig:heatmap}, where we report the similarity between the task keys, used to select the parameters during inference. The heatmap shows how the similarity is indeed higher for tasks presented in order (e.g., task 1 with task 2, task 3 with task 4, etc.). The fact that the similarity between tasks/keys is reflected in the parameters space (Figure~\ref{fig:tsne}), suggests that future work may use directly the parameters and the network's activations to estimate the task ID during inference (e.g., by comparing normalization statistics).

\section{Conclusion}
In this paper, we present Continual LayerNorm, a new method for rehearsal-free continual learning.
Contrary to prompt-based approaches, we propose a simple strategy and tune the LayerNorms parameters of the pre-trained Vision Transformer.
Moreover, we present a single-stage variant to halve the inference time and a training strategy to improve the classifier robustness during inference.
Experiments on CIFAR-100 and ImageNet-R show that
our method either outperforms or is on par with the current state-of-the-art while being computationally cheaper. 

\vspace{5pt}
\noindent\textbf{Acknowledgements.}  
This work was supported by the MUR PNRR project FAIR - Future AI Research (PE00000013) funded by the NextGenerationEU, the PRIN project LEGO-AI (Prot. 2020TA3K9N), and the EU H2020 AI4Media No. 951911 project. It was carried out in the Vision and Learning joint laboratory of FBK and UNITN. 

{\small
\bibliographystyle{ieee_fullname}
\bibliography{egbib}
}

\end{document}